\documentclass{article}

\usepackage{arxiv}

\usepackage[utf8]{inputenc} 
\usepackage[T1]{fontenc}    
\usepackage{hyperref}       
\usepackage{url}            
\usepackage{booktabs}       
\usepackage{amsfonts}       
\usepackage{nicefrac}       
\usepackage{microtype}     
\usepackage{lipsum}		
\usepackage{graphicx}
\usepackage{natbib}
\usepackage{doi}
\usepackage{subcaption}
\usepackage{placeins}
\usepackage{makecell}
\usepackage{multirow}
\usepackage[ruled,vlined]{algorithm2e}
\usepackage{amsmath}
\usepackage{amssymb}
\usepackage[capitalize]{cleveref}
\newcommand{\eg}{\textit{e.g.}}

\title{Discriminative Flow Matching via Local Generative Predictors}

\author{Om Govind Jha \\
    Indian Institute of Science Education and Research \\
    Bhopal, India \\
    \texttt{om22@iiserb.ac.in} \\
    \And
    Manoj Bamniya \\
    Indian Institute of Technology \\
    Guwahati, India \\
    \texttt{manojbamniya832@gmail.com} \\
    \And
    Ayon Borthakur \\
    Indian Institute of Technology \\
    Guwahati, India \\
    \texttt{ayon.borthakur@iitg.ac.in}
}

\renewcommand{\shorttitle}{Discriminative Flow Matching via Local Generative Predictors}

\hypersetup{
pdftitle={Discriminative Flow Matching via Local Generative Predictors},
pdfsubject={cs.CV, cs.LG},
pdfauthor={Om Govind Jha, Manoj Bamniya, Ayon Borthakur},
pdfkeywords={Transformer, Flow Matching, Image Classification, Object Detection, Representation Learning},
}
\begin{document}
\maketitle

\begin{abstract}
  Traditional discriminative computer vision relies predominantly on static projections, mapping input features to outputs in a single computational step. Although efficient, this paradigm lacks the iterative refinement and robustness inherent in biological vision and modern generative modelling. In this paper, we propose Discriminative Flow Matching, a framework that reformulates classification and object detection as a conditional transport process. By learning a vector field that continuously transports samples from a simple noise distribution toward a task-aligned target manifold - such as class embeddings or bounding box coordinates - we are at the interface between generative and discriminative learning. Our method attaches multiple independent flow predictors to a shared backbone. These predictors are trained using local flow matching objectives, where gradients are computed independently for each block. We formulate this approach for standard image classification and extend it to the complex task of object detection, where targets are high-dimensional and spatially distributed. This architecture provides the flexibility to update blocks either sequentially to minimise activation memory or in parallel to suit different hardware constraints. By aggregating the predictions from these independent flow predictors, our framework enables robust, generative-inspired inference across diverse architectures, including CNNs and vision transformers.
  \keywords{Transformer \and Flow Matching \and Image Classification \and \ Object Detection \and Representation Learning}
\end{abstract}

\section{Introduction}
For the past few years, the dominant paradigm in discriminative computer vision has been \textit{static projection}. Architectures such as ResNets [[\cite{he2016deep}] and vision transformers (ViTs) [[\cite{dosovitskiy2020image}] function as fixed mappings that project input features directly onto a set of logits or coordinates in a single computational step. While highly optimised, this static approach stands in contrast to biological vision and modern generative modelling, both of which rely on dynamic, iterative refinement to resolve ambiguity and generate high-fidelity structures.

The success of Denoising Diffusion Probabilistic Models (DDPMs) [\cite{ho2020denoising} has sparked interest in bringing these "dynamic" capabilities to discriminative tasks. Recent works like DiffusionDet [[\cite{chen2023diffusiondet}] have demonstrated that object detection can be formulated as a denoising process, where random boxes are progressively refined into precise detections. More recently, Flow Matching [\cite{lipman2023flow}] has emerged as an alternative to diffusion, offering simulation-free training and straighter transport trajectories. However, adapting these continuous-time processes to discriminative tasks introduces computational bottlenecks. Training a deep network to approximate a global vector field typically requires backpropagating through ODE solvers[[\cite{Shah2018BPODE}] or massive unrolled graphs, causing memory usage to scale linearly with depth and making optimisation notoriously unstable.

To mitigate the costs of global backpropagation (BP), the \textit{Local Learning} alternatives have been explored, such as the Forward-Forward (FF) algorithm [\cite{hinton2022forward}]. While FF and its variants (\eg, CFF [\cite{aghagolzadeh2025contrastive}], SFF [\cite{zhu2025stochastic}]) effectively decouple layer training, they have struggled to scale beyond global image classification. Their reliance on scalar ``goodness'' objectives makes extending them to dense prediction tasks challenging [\cite{zhao2019object}], as they struggle to preserve the high-dimensional spatial information required for precise localisation. A closely related approach, NoProp [\cite{li2025noprop}], takes a more principled step by eliminating both forward and backward propagation entirely: each block independently learns to denoise a noisy version of the target label embedding, drawing directly on diffusion and flow-matching theory. While NoProp represents a compelling advance in credit-assignment-free learning and demonstrates competitive accuracy on image classification benchmarks, it is designed around fixed, pre-noised label targets broadcast uniformly to every block. This architectural choice, while effective for classification, does not naturally extend to structured output spaces where targets are high-dimensional and spatially distributed, precisely the setting our work addresses. Our approach follows the spirit of NoProp by using block-local flow matching objectives without requiring full end-to-end backpropagation through the diffusion steps.
In this paper, we propose \textbf{Discriminative Flow Matching}, a framework that leverages the representational strength of continuous flows to build robust, generative-inspired classifiers and detectors. Flow Matching learns a vector field that continuously transports samples from a source distribution toward a target, offering a smooth, tractable alternative to diffusion-based generation. Instead of treating the network with a single static projection head, we conceptualise inference as a consensus-driven transport process. We use an ensemble of independent flow predictors, where multiple parallel heads learn distinct local vector fields to transport a noisy latent state toward a task-aligned target manifold - whether a class embedding or a set of bounding box coordinates. By aggregating these independent generative trajectories, our method achieves highly robust single-step inference, effectively bridging the gap between generative continuous-time dynamics and fast, accurate discrimination.

Our contributions are as follows:

\begin{enumerate}

\item We extend conditional flow matching to discriminative vision tasks,
showing that both image classification and object detection can be formulated
as transporting Gaussian noise to task-aligned target manifolds, such as class
embeddings and sets of bounding box coordinates.

\item We introduce a \textbf{Generative Flow Ensemble} that departs from
standard static projection. Instead of training multiple independent models as
in traditional ensembles, we attach multiple vector-field predictors to a
shared backbone. Aggregating their predictions provides a robust consensus for
task inference while avoiding the memory and computational cost of training
multiple full networks.

\item We demonstrate experimentally that this framework scales to both image
classification and object detection across CNN and Vision Transformer (ViT)
architectures, with models trained both from scratch and using pretrained
backbones, achieving competitive accuracy while providing a generative
interpretation of discriminative inference.

\item We provide an empirical analysis of generative inference dynamics across
architectures. Our results show that aggregating predictions from multiple flow
predictors enable robust single-step ensemble inference, while multi-step ODE
integration can become unstable when predictors are not specialised for
specific time intervals, leading to trajectory drift, particularly in
convolutional architectures [\cite{chen2018neuralode,lipman2023flow}].

\end{enumerate}

\section{Related Work}

\subsubsection{Flow Matching and Neural ODEs.}
Flow Matching [\cite{lipman2023flow}] was introduced as a scalable method for training Continuous Normalising Flows (CNFs) [\cite{chen2018neuralode}] by regressing a vector field directly onto a conditional probability path. While it is primarily dominant in generative tasks (\eg, image synthesis), its utility for discrimination remains underexplored. Closely related to our work is \textit{NoProp} [\cite{li2025noprop}], from which we draw inspiration. NoProp demonstrates that classification can be achieved without global forward or backward propagation by having independent network blocks learn to denoise targets using localised flow matching objectives. 

While we share the core motivation of utilising decoupled generative objectives to circumvent global optimisation bottlenecks, we distinguish our work in two key dimensions. First, we extend the formulation beyond global classification to the more complex task of \textbf{Object Detection}, which requires transporting noise to permutation-invariant sets of bounding box coordinates rather than static class vectors. Secondly, while NoProp largely decouples learning from traditional hierarchical feature extraction, we propose a Generative Flow Ensemble. By attaching independent flow predictors to a shared, hierarchical backbone (\eg, ViT or ResNet), we leverage rich global features to achieve highly robust, single-step ensemble inference, offering a practical bridge between generative continuous-time dynamics and standard discriminative architectures.

\subsubsection{Dynamic and Generative Recognition.}
The paradigm of treating perception as a dynamic process is rapidly evolving. \textit{DiffusionDet} [\cite{chen2023diffusiondet}] pioneered the use of generative diffusion for object detection, treating bounding boxes as signals to be denoised. While effective, the stochastic nature of diffusion often leads to slow inference. \textit{DeFloMat} [\cite{deflo_anonymous}] recently adapted Flow Matching to this domain to accelerate sampling. However, these methods rely on heavy end-to-end backpropagation through the entire backbone and decoder. This global dependency imposes strict limits on model depth and batch size. Our work uses the Optimal Transport paths of Flow Matching [\cite {lipman2023flow}], but avoids the global gradient lock. By training flow dynamics in decoupled blocks, we retain the benefits of iterative refinement without the prohibitive memory cost.

\subsubsection{Efficient Learning.}
Standard Backpropagation (BP) imposes strict memory and scaling limits due to the need to store intermediate activations across the entire network graph. To circumvent this, \textit{Forward-Forward (FF)} [\cite{hinton2022forward}] proposed replacing the backward pass with a second forward pass of "negative data", optimising a local goodness function layer-by-layer.

While these efficient learning methods are promising, and further strengthened by contrastive formulations such as 
\textit{CFF} [\cite{aghagolzadeh2025contrastive}] and \textit{SFF} 
[\cite{zhu2025stochastic}], these local methods are primarily designed for 
classification settings. Extending them to dense prediction tasks introduces 
additional challenges: scalar objective functions compress the rich, 
high-dimensional error signals required for precise spatial localisation, and 
layer-wise updates may not always produce the globally coherent representations 
that structured outputs demand.

Flow Matching offers a complementary alternative for such settings. Instead of 
optimising a scalar goodness score, it learns a continuous vector field that 
transports latent representations toward structured targets. This provides a 
dense, directional supervision signal that is naturally suited to the spatial and structural complexity of dense prediction.

Rather than discarding hierarchical backpropagation entirely, our Discriminative Flow Matching maintains a shared backbone for robust feature extraction but decouples the generative prediction stage. By regressing a high-dimensional vector field, we provide the dense, directional supervision needed for complex tasks such as object detection. Furthermore, by isolating gradient computation to independent flow heads, we achieve robust multi-trajectory generative inference without the massive memory overhead associated with standard global ensembles or unrolled continuous-time ODEs.

\section{Methods}

In this section, we present our block-wise local training paradigm based on flow matching for object detection and image classification. 
\subsection{Problem Formulation}

\subsubsection{Flow Matching in Embedding Space}

Our approach learns to transport samples from a simple noise distribution to a structured target embedding through a learned vector field. We formulate this as a continuous-time flow matching problem.

\textbf{Initial State.} We begin with samples from a standard Gaussian distribution:
\begin{equation}
z_0 \sim \mathcal{N}(0, I)
\end{equation}

\textbf{Target State.} The target state $z_1$ is a learnable, deterministic embedding conditioned on the ground truth label or detection:
\begin{itemize}
    \item \textit{For Classification:} $z_1 = W_{\text{embed}}[y]$, where $W_{\text{embed}} \in \mathbb{R}^{C \times d}$ is a learnable class embedding matrix for $C$ classes and embedding dimension $d$.
    \item \textit{For Object Detection:} $z_1$ is a set of $M$ target queries, where each query $j$ is constructed as:
    \begin{equation}
    z_{1_j} = W_{\text{class}}[c_j] + W_{\text{box}}b_j
    \end{equation}
    where $W_{\text{class}} \in \mathbb{R}^{(C+1) \times d}$ includes a background class, and $W_{\text{box}} \in \mathbb{R}^{d \times 4}$ projects bounding box coordinates $b_j \in [0,1]^4$.
\end{itemize}

\textbf{Probability Path.} We define a linear interpolation path between noise and target[\cite{lipman2023flow,peyre2019ot}]:
\begin{equation}
z_t = (1-t)z_0 + tz_1, \quad t \in [0, 1]
\end{equation}

\textbf{Target Vector Field.} The derivative of this path defines the direction the network must learn:
\begin{equation}
v_t = \frac{\mathrm{d}}{\mathrm{d}t}z_t = z_1 - z_0
\end{equation}

\subsection{Block-wise Local Training Architecture}

\subsubsection{Multi-Block Decomposition}

Instead of constructing a single deep sequential network, we decompose the model
into $T$ parallel vector-field predictors
$\{v_{\theta_1}, v_{\theta_2}, \ldots, v_{\theta_T}\}$.
Each predictor $v_{\theta_k}$ is a neural module with parameters $\theta_k$
that shares the same architecture across blocks and receives identical inputs,
namely the current latent state $z_t$, image features $f(x)$ extracted from the
shared backbone, and the time variable $t$.
Each block predicts the same target vector field $z_1 - z_0$ and is optimised
using its own local loss.
Although these predictors are trained independently and without a fixed temporal assignment, they can be applied sequentially or aggregated at inference time to approximate the underlying transport dynamics.

\subsubsection{Vector Field Prediction}

Each block predicts the vector field using the following architecture.

\textbf{Feature Extraction.} Image features are extracted using a shared backbone:
\begin{equation}
f(x) = \text{Backbone}(x)
\end{equation}
Unlike local learning approaches such as Stochastic Forward-Forward [\cite{zhu2025stochastic}] and NoProp [\cite{li2025noprop}], which rely on layer-wise training with fixed or non-jointly optimised intermediate representations, our method continuously updates a shared backbone. To ensure memory efficiency, we adopt a sequential gradient accumulation strategy widely utilised in Multi-Task Learning (MTL) optimisation [\cite{vandenhende2021multi, yu2020gradient}]. By backpropagating through the parallel flow heads sequentially and accumulating their gradients into the shared backbone, the network learns a robust consensus representation while strictly bounding the peak activation memory to the backbone plus a single head at any given time. For image classification, we evaluate both a custom multi-stage Convolutional Neural Network (CNN) and a Vision Transformer (ViT-Tiny variant) that outputs $f(x) \in \mathbb{R}^d$, while for object detection, we use a pre-trained ResNet-50 with features projected to dimension $d$. The configuration details are provided in the supplementary file.

\textbf{Time Embedding.} The scalar time variable $t \in [0,1]$ is embedded into a dense vector space using a two-layer Multi-Layer Perceptron (MLP):
\begin{equation}
e_t = \text{TimeEmbed}(t) \in \mathbb{R}^d
\end{equation}
This temporal conditioning is a crucial mechanism for the continuous generative process [\cite{ho2020denoising, lipman2023flow}]. Because the underlying vector field evolves continuously from the source noise distribution at $t=0$ to the target data manifold at $t=1$, the time embedding informs the network of the current integration step and allows the shared weights within the flow blocks to dynamically adjust their velocity predictions depending on the position of the latent state along the trajectory [\cite{lipman2023flow}].

\textbf{Vector Field Architecture.} For classification, the time embedding is first added to the latent state to provide temporal conditioning. The conditioned state is then concatenated with the image features and processed by a Multi-Layer Perceptron (MLP) to predict the vector field:
\begin{align}
h &= \text{Concat}(f(x), z_t + e_t) \\
v_{\theta_k}(z_t, f(x), t) &= \text{MLP}_k(h) \in \mathbb{R}^d
\end{align}
For object detection, the blocks instead use transformer layers with self-attention and cross-attention mechanisms to process query embeddings conditioned on image features and time. We describe the local training objectives in the supplementary. 

\subsection{Training Procedure}

\begin{figure}[t]
    \centering
   
    \includegraphics[width=\textwidth]{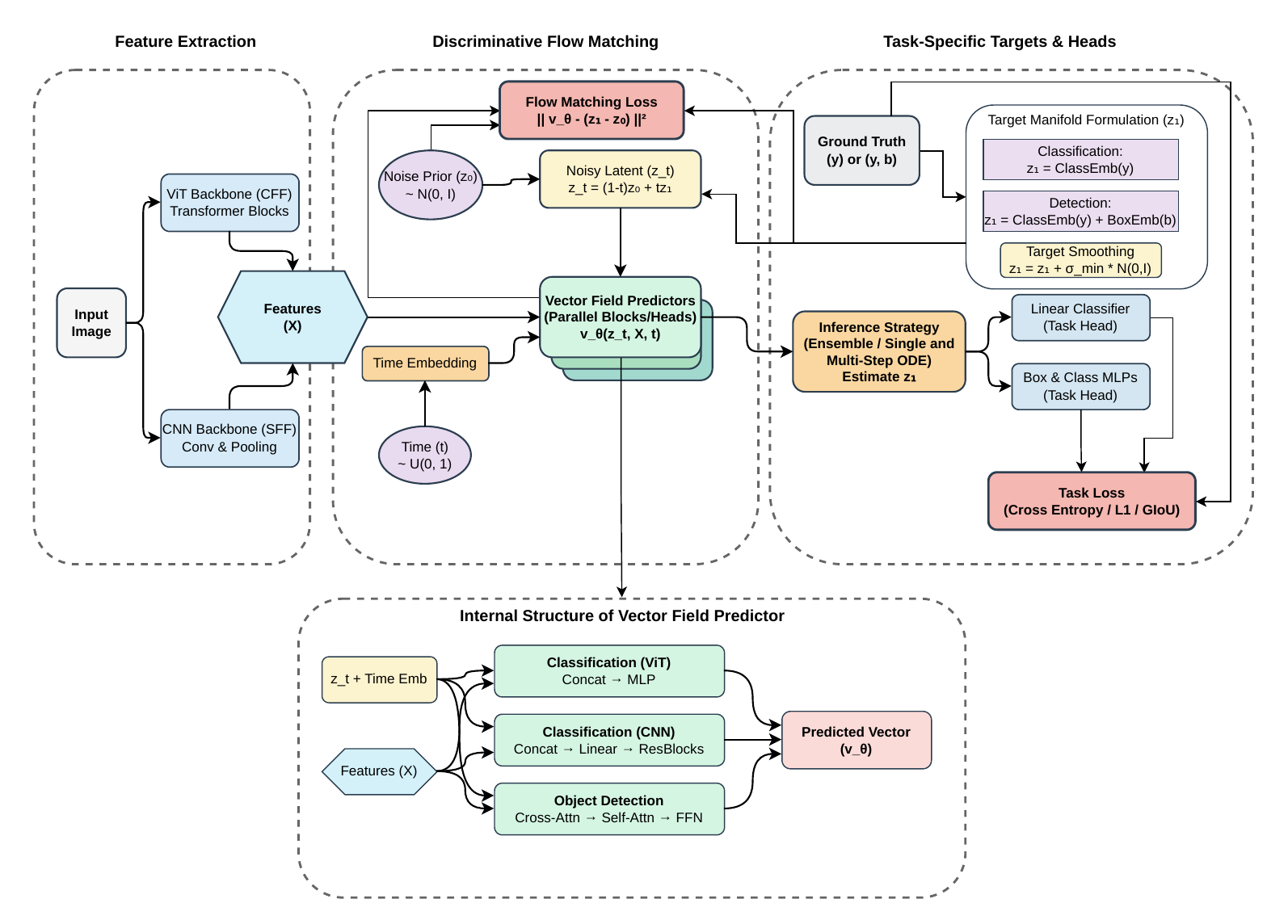}
    \caption{Overview of the proposed Discriminative Flow Matching framework}
    \label{fig:flow_matching_arch}
\end{figure}

Our training procedure processes each block sequentially, and later we also compare it with a parallel training approach. For each epoch, we iterate over all $T$ blocks and process the entire dataset for each block. ~\cref{fig:flow_matching_arch} depicts this Flow Matching Framework. ~\cref{alg:training} presents the complete training procedure. 

\begin{algorithm}[t]
\small
\caption{Block-wise Local Training for Flow Matching}
\label{alg:training}
\KwIn{Dataset $\mathcal{D}$, number of blocks $T$, epochs $E$}
\KwIn{Initialized backbone $f(\cdot)$, blocks $\{v_{\theta_k}\}_{k=1}^{T}$, embeddings}
\KwOut{Updated block parameters $\{\theta_k\}$, backbone, and embeddings}

\For{epoch $e = 1$ to $E$}{
  \For{block $k = 1$ to $T$}{
    \For{mini-batch $(x, y) \in \mathcal{D}$}{

      \tcc{Sample latent states}
      $z_1 \gets \text{CreateTargetEmbedding}(y)$\;
      $z_0 \sim \mathcal{N}(0, I)$\;
      $t \sim \mathcal{U}[0,1]$\;
      $z_t \gets (1 - t) z_0 + t z_1$\;

      \tcc{Forward pass}
      $f(x) \gets \text{Backbone}(x)$\;
      $\hat{v}_t \gets v_{\theta_k}(z_t, f(x), t)$\;

      \tcc{Flow matching loss}
      $\mathcal{L}_{\mathrm{flow}}^{(k)} \gets
      \left\| \hat{v}_t - (z_1 - z_0) \right\|_2^2$\;

      \tcc{Auxiliary task supervision (training only)}
      $\tilde{z}_1 \gets z_t + (1 - t)\hat{v}_t$\;
      $\mathcal{L}_{\mathrm{task}}^{(k)} \gets
      \text{TaskLoss}(\tilde{z}_1, y)$\;

      \tcc{Total loss}
      $\mathcal{L}_{\mathrm{total}}^{(k)} \gets
      \mathcal{L}_{\mathrm{flow}}^{(k)} + \lambda \mathcal{L}_{\mathrm{task}}^{(k)}$\;

      \tcc{Backward pass (local to block $k$)}
      Compute $\nabla_{\theta_k} \mathcal{L}_{\mathrm{total}}^{(k)}$
      and $\nabla_{\text{backbone}} \mathcal{L}_{\mathrm{total}}^{(k)}$\;
      Update $\theta_k$, backbone, and embeddings\;

      \tcc{No gradients are propagated to other blocks $\theta_j$, $j \neq k$}
    }
  }
}
\end{algorithm}

\textbf{Key Property - Gradient Locality:} The crucial distinction from standard backpropagation is that the gradient computation for block $k$ is independent:
\begin{equation}
\nabla_{\theta_k} \mathcal{L}_{\text{total}}^{(k)} \quad \text{does not depend on } \theta_j \text{ for } j \neq k
\end{equation}
The computational graph only includes: Backbone $\rightarrow$ Block $k$ $\rightarrow$ Loss. This enables efficient memory usage by storing activations for only one block at a time, rather than maintaining the full computation graph across all $T$ blocks.

For details, refer to the memory efficiency analysis section in the supplementary.

\subsection{Inference Strategies for Learned Flow Dynamics}

We evaluate the learned representations using four inference strategies:
Linear Probe, Single-Step, Ensemble, and
Multi-Step (ODE) inference. The linear probe evaluates the
quality of backbone representations by training a classifier on frozen
features [\cite{alain2016understanding}]. Single-step inference estimates the target using one Euler
update $\hat{z}_1 = z_0 + v_\theta(z_0,0)$. Ensemble inference aggregates
predictions from multiple flow heads, while multi-step inference solves
the corresponding ODE using an Euler solver.
Further implementation details are provided in the supplementary.

\section{Propositions}

We now provide a theoretical analysis of the proposed flow-matching framework, addressing convergence, ensemble behaviour, and stability in the practical setting of our implementation, where the backbone is shared and jointly updated across all blocks. Please refer to the supplementary file for a detailed analysis of the propositions.

\subsection{Convergence Analysis}

\textbf{Proposition 1 (Local Consistency of Euler Integration).}
Let $v_\theta(z,t)$ denote the learned vector field obtained via flow matching, and
$v^*(z,t)$ the ground-truth vector field induced by the linear interpolation path
between the initial noise $z_0$ and the target embedding $z_1^*$.
Assume that $v_\theta$ is $L$-Lipschitz in $z$ and satisfies the local flow-matching
error bound
\begin{equation}
\mathbb{E}_{z_0,z_1,t}\!\left[\|v_\theta(z_t,t)-v^*(z_t,t)\|_2^2\right] \le \epsilon .
\end{equation}
Then, for a sufficiently small Euler step size $\Delta t$, the discrete Euler update
\begin{equation}
z_{i+1} = z_i + \Delta t\, v_\theta(z_i,t_i)
\end{equation}
satisfies the following local consistency bound:
\begin{equation}
\mathbb{E}\!\left[\|z_{i+1}-z_1^*\|_2^2\right]
\le
\mathbb{E}\!\left[\|z_i-z_1^*\|_2^2\right]
+ \mathcal{O}(\Delta t\,\epsilon)
+ \mathcal{O}(\Delta t^2).
\end{equation}

\subsection{Trajectory Refinement Analysis}

\textbf{Proposition 2 (Progressive Trajectory Refinement).}
Let the continuous-time vector field $v^*(z,t)$ be approximated by a sequence of
$T$ learned vector fields $\{v_{\theta_k}\}_{k=1}^T$, each applied over a disjoint
time interval $[t_k, t_{k+1}]$ with step size $\Delta t = 1/T$.
Assume that each block $v_{\theta_k}$ satisfies a local approximation error
\[
\mathbb{E}\!\left[\|v_{\theta_k}(z,t) - v^*(z,t)\|_2^2\right] \le \epsilon_k,
\quad \forall t \in [t_k, t_{k+1}].
\]
Then the accumulated integration error after $K \le T$ steps satisfies
\[
\mathbb{E}\!\left[\|z_K - z_K^*\|_2^2\right]
\le
\sum_{k=1}^K \mathcal{O}(\Delta t \, \epsilon_k) + \mathcal{O}(K \Delta t^2),
\]
where $z_K^*$ denotes the ideal continuous-time trajectory evaluated at time $t_K$.
In particular, if the local errors $\epsilon_k$ decrease across blocks,
the discrete trajectory exhibits monotonic refinement toward the task-aligned manifold.

\subsection{Training Stability Guarantees}

\textbf{Proposition 3 (Bounded Gradient Variance with a Shared Backbone).}
Let $\phi$ denote the parameters of a backbone network shared across $T$ flow-matching
blocks $\{\theta_k\}_{k=1}^T$, and suppose all parameters are trained jointly.
Assume that for each block $k$, the total loss is given by
\[
\mathcal{L}^{(k)} = \mathcal{L}_{\mathrm{flow}}^{(k)} + \lambda \, \mathcal{L}_{\mathrm{task}}^{(k)} .
\]
Then the variance of the gradient with respect to the block parameters $\theta_k$
satisfies
\begin{equation}
\mathrm{Var}\!\left[\nabla_{\theta_k}\mathcal{L}^{(k)}\right]
\;\le\;
\sigma_{\mathrm{flow}}^2
+ \lambda^2 \sigma_{\mathrm{task}}^2
+ \beta^2 \, \mathrm{Var}\!\left[\nabla_{\phi}\mathcal{L}_{\mathrm{shared}}\right],
\end{equation}
where $\sigma_{\mathrm{flow}}^2$ and $\sigma_{\mathrm{task}}^2$ denote the intrinsic
gradient variances of the flow-matching and task losses, respectively, and
$\beta$ captures the sensitivity of block gradients to backbone updates.
Under these conditions, the gradient variance of each block remains bounded
and does not grow with the number of blocks $T$.

\subsection{Memory Complexity Analysis}

\textbf{Proposition 4 (Constant Activation Memory).}
\emph{Under blockwise local training, the peak activation memory of our method is
$\mathcal{O}(B \cdot d)$ and is independent of the number of blocks $T$, where $B$
denotes the batch size, and $d$ denotes the latent dimension.}

\section{Experiments}

In this section, we present the experiments to evaluate our proposed method. We analyse performance on object detection and image classification tasks, as well as memory consumption during training.

\subsection{Object Detection}

\begin{figure*}[t]
    \centering
    \begin{subfigure}{0.49\textwidth}
        \centering
        \includegraphics[width=\linewidth]{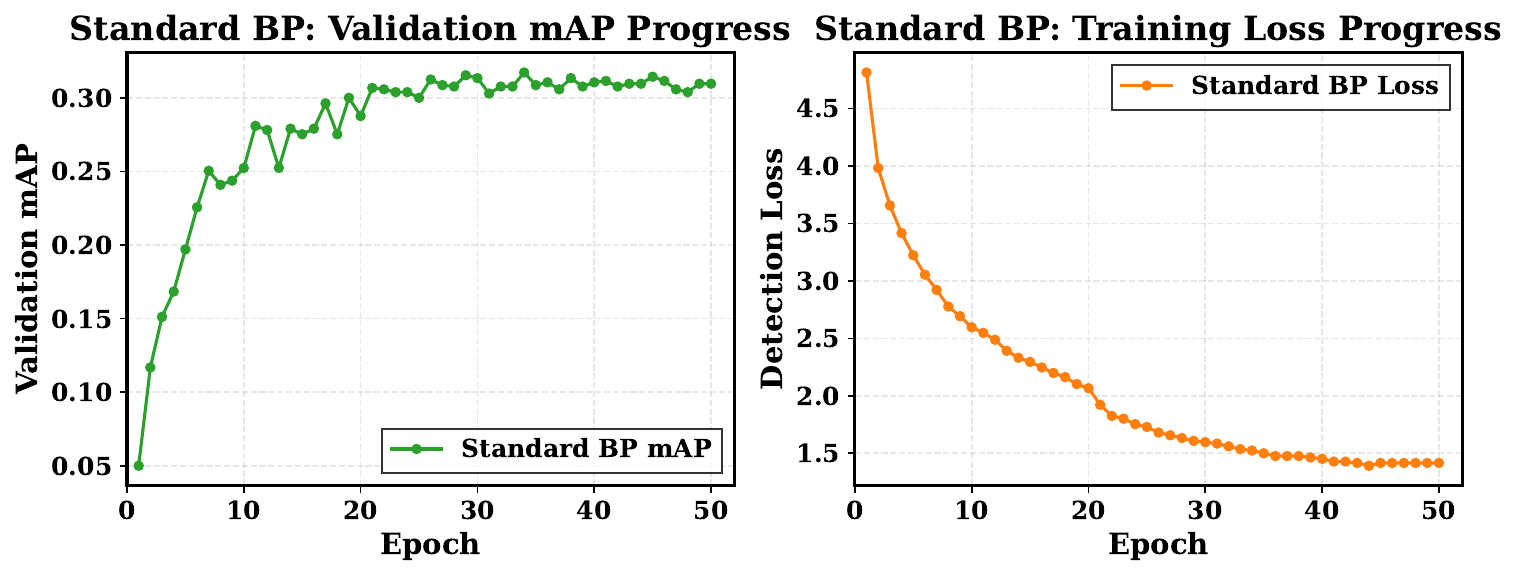}
        \caption{Standard Backpropagation Training}
        \label{fig:sub_backprop}
    \end{subfigure}
    \hfill
    \begin{subfigure}{0.49\textwidth}
        \centering
        \includegraphics[width=\linewidth]{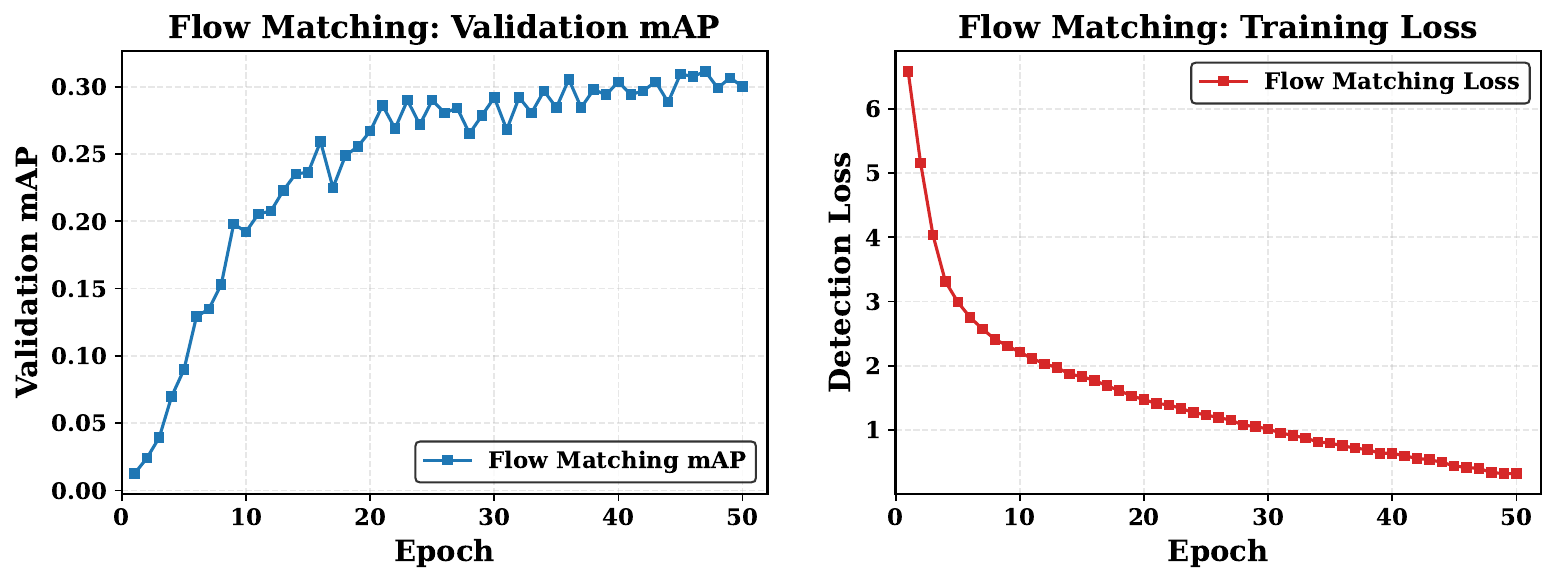}
        \caption{Flow Matching Training}
        \label{fig:sub_flow}
    \end{subfigure}
    \caption{Training progress comparison on PASCAL VOC using ResNet-50. Both validation mAP and training loss are shown over 50 epochs.}
    \label{fig:pascal_voc_normal}
\end{figure*}

We evaluate performance on the PASCAL VOC dataset using a ResNet-50 backbone. ~\cref{tab:object_detection_summary} summarises the key performance metrics for these experiments. ~\cref{fig:pascal_voc_normal} compares the training progress of a standard backpropagation model against a model trained using Flow Matching. The plots show that both methods achieve a similar Mean Average Precision (mAP) of around 30 mAP, though the convergence behaviour of the training loss differs.

To simulate a more constrained training environment, we repeat the experiment with a batch size of 1. Training is less stable under these conditions for both methods, which is expected. The plot for this is given in the supplementary.

\begin{table}[t]
    \centering
    \caption{Performance summary for object detection tasks. Performance is peak validation mAP over 50 epochs (unless noted).}
    \label{tab:object_detection_summary}
    \resizebox{0.9\textwidth}{!}{ 
        \begin{tabular}{l l c c c c}
            \toprule
            \textbf{Method} & \textbf{Dataset} & \textbf{Performance (mAP)} & \textbf{Memory (MB)} & \makecell{\textbf{Layers after} \\ \textbf{Backbone}} \\
            \midrule
            Backpropagation (BS=8) & PASCAL VOC & 31 & 1274.84 & 4 (Decoder) \\
            Flow Matching (BS=8) & PASCAL VOC & 30 & 1261.14 & 4 (Decoder)  \\
            \midrule
            Backpropagation (BS=1) & PASCAL VOC & 23 & 579.23 & 4 (Decoder) \\
            Flow Matching (BS=1) & PASCAL VOC & 24 & 510.84 & 4 (Decoder)  \\
            
            \bottomrule
        \end{tabular}
    }
\end{table}

\subsection{Memory Consumption Analysis}

\begin{figure}[h]
    \centering
    \includegraphics[width=0.7\linewidth]{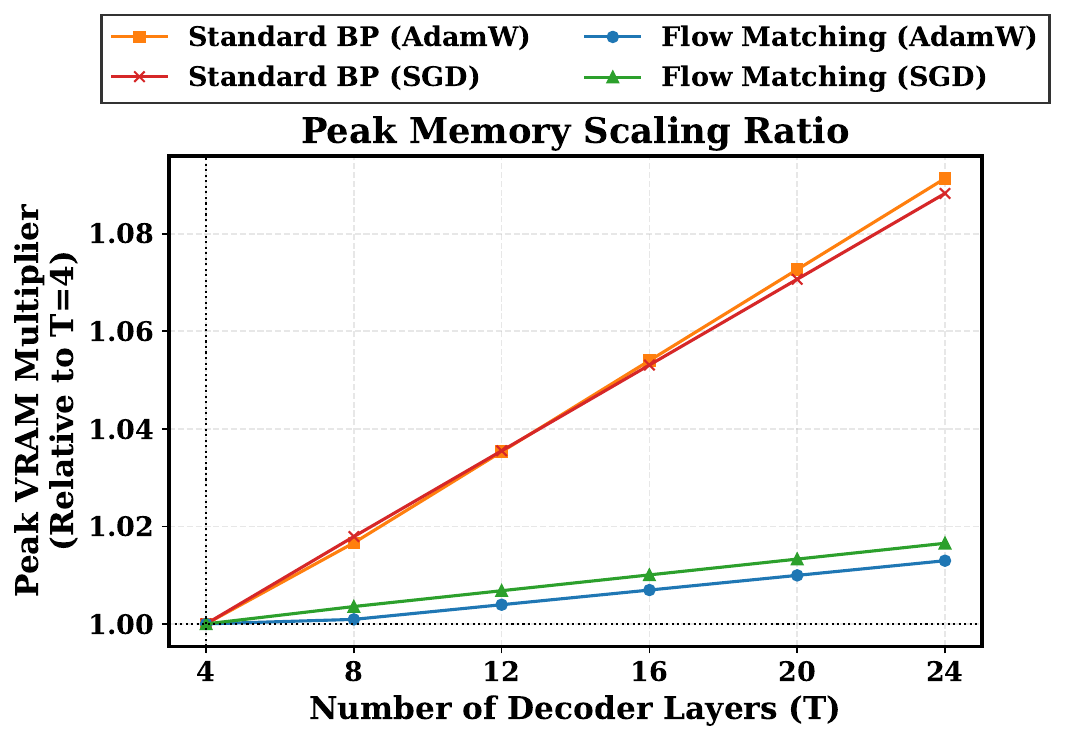}
    \caption{Memory consumed (in MB) versus the number of decoders for FM and Backprop methods on Pascal Object Detection Tasks for batch size 8}
    \label{fig:memory}
\end{figure}

 We measured the memory consumed by our method versus standard Backprop as a function of the number of decoder layers in the model. As illustrated in ~\cref{fig:memory} and in ~\cref{tab:classification_summary}, our method consumes less memory, and this advantage increases as the model depth increases beyond the backbone. Both models in ~\cref{fig:memory} use the same ResNet-50 backbone with Batch Normalisation.
 The reduced memory footprint arises from training each block using a locally defined objective and constructing smaller computation graphs per update. While end-to-end backpropagation can also be made memory-efficient through techniques such as checkpointing or staged training, in our formulation, the memory reduction follows directly from the decomposed optimisation strategy rather than any additional modifications as required in standard backpropagation.

\subsection{ViT Fine-Tuning Task}

\begin{figure}[t]
    \centering
    \includegraphics[width=\linewidth]{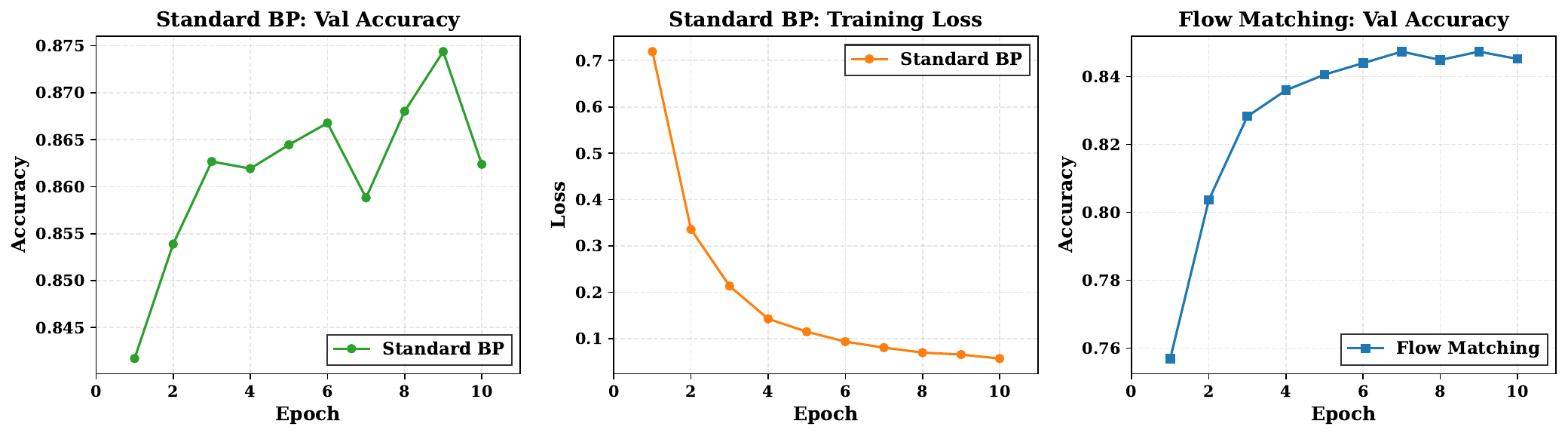}
    \caption{Fine-tuning progress comparison on CIFAR-100 using a ViT model.
Standard Backpropagation (left and center) is compared against Flow Matching (right) over 10 epochs.}
    \label{fig:cifar100_finetuning}
\end{figure}

We compare the fine-tuning performance of a standard Vision Transformer (ViT) with our Flow Matching method and backpropagation across CIFAR-10, CIFAR-100, and ImageNet-R. ~\cref{tab:classification_summary} reports the peak Top-1 accuracy after 10 epochs, together with GPU memory usage and architectural configuration. ~\cref{fig:cifar100_finetuning} describes the fine-tuning profiles using standard backpropagation and Flow Matching. 

On CIFAR-100, Flow Matching achieves 84.47\% accuracy, which is very close to the standard fine-tuning accuracy of 86.24\%. Although the standard approach performs slightly better, Flow Matching reduces memory consumption, indicating a modest efficiency gain with a small performance trade-off. On CIFAR-10, Flow Matching achieves 98.19\% accuracy compared to 96.74\% for standard fine-tuning, while also requiring less memory. This suggests that Flow Matching can be both more efficient and more effective on simpler datasets. On the high-resolution and complex ImageNet-R dataset, Flow Matching achieves 73.4\% (vs. 75.9\% for fine-tuning). Similar to CIFAR-100, the standard approach achieves slightly higher accuracy, while Flow Matching uses less memory.

Across all experiments, both methods use the same backbone and identical post-backbone architecture (four ViT head layers), ensuring that performance differences arise from the optimisation strategy rather than architectural changes. Overall, the results show that Flow Matching can match or exceed standard fine-tuning on simpler datasets such as CIFAR-10 while consistently reducing memory usage, although it may incur a small accuracy gap on more complex datasets. In addition to fine-tuning, we report results from scratch training in the supplementary. 

\begin{table}[t]
    \centering
    \caption{Fine-Tuning performance summary for classification tasks. Performance is peak Top-1 Accuracy after 10 epochs (unless noted).}
    \label{tab:classification_summary}
    \small
    \setlength{\abovecaptionskip}{4pt}
    \setlength{\belowcaptionskip}{-2pt}
    \renewcommand{\arraystretch}{0.92}
    \begin{tabular}{l l c c c c}
        \toprule
        \textbf{Method} & \textbf{Dataset} & \textbf{Performance (Acc@1)} & \textbf{Memory (MB)} & \makecell{\textbf{Layers after} \\ \textbf{Backbone}} \\
        \midrule
        Standard Fine-Tune & CIFAR-100 & 0.8624 & 5292.55 & 4 (ViT Head) \\
        Flow Matching & CIFAR-100 & 0.8447 & 5082.63 & 4 (ViT Head)  \\
        \midrule
        Standard Fine-Tune & CIFAR-10 & 0.9674 & 5291.49 & 4 (ViT Head) \\
        
        Flow Matching & CIFAR-10 & 0.9819 & 5081.67 & 4 (ViT Head) \\
        \midrule
        Standard Fine-Tune & ImageNet-R & 0.759 & 9170.04 & 4 (ViT Head) \\
        Flow Matching & Imagenet-R & 0.724 & 8959.83 & 4 (ViT Head) \\
        \bottomrule
    \end{tabular}
\end{table}

\subsection{Analysis of Inference Dynamics}
We conducted an experiment to understand the effects of architecture (CNN vs. ViT) and training modality (Parallel vs. Sequential) on Flow Matching performance. Our experiments on CIFAR-10 and CIFAR-100 reveal distinct inductive biases that dictate the optimal training strategy for each architecture. The results of the experiment are in ~\cref{tab:inference_ablation}.

\begin{table}[t]
\centering
\caption{Inference types and performance of ViT and CNN.}
\label{tab:inference_ablation}
\resizebox{\textwidth}{!}{%
\begin{tabular}{|l|l|c|c|c|c|}
\hline
\textbf{Dataset} & \textbf{Model (Arch + Mode)} & \textbf{Feature Quality} & \multicolumn{3}{c|}{\textbf{Generative Inference Accuracy (\%)}} \\ \cline{3-6}
 & & \textbf{Linear Probe} & \textbf{Single-Step} & \textbf{Ensemble} & \textbf{Multi-Step} \\ \hline

\multirow{4}{*}{\textbf{CIFAR-10}}
 & CNN (Parallel) & 80.01\% & 77.26\% & 77.54\% &  78.56\% \\ \cline{2-6}
 & CNN (Sequential) & 82.73\% & 69.61\% & 75.52\% & 70.28\% \\ \cline{2-6}
 & ViT (Parallel) & 64.21\% & 60.10\% & 60.46\% & 59.80\% \\ \cline{2-6}
 & ViT (Sequential) & 82.02\% & 80.10\% & 80.05\% & 80.10\% \\ \hline

\multirow{4}{*}{\textbf{CIFAR-100}}
 & CNN (Parallel) & 82.52\% & 57.86\% & 59.20\% & 34.76\% \\ \cline{2-6}
 & CNN (Sequential) & 85.13\% & 14.65\% & 15.45\% & 34.09\% \\ \cline{2-6}
 & ViT (Parallel) & 88.71\% & 78.99\% & 80.67\% & 67.77\% \\ \cline{2-6}
 & ViT (Sequential) & 79.84\% & 75.37\% & 75.50\% & 74.25\% \\ \hline

\end{tabular}%
}
\end{table}

\noindent\textbf{Feature Quality.}
Sequential training improves feature quality for CNN backbones, as reflected by higher linear probe accuracy (85.13\% vs.\ 82.52\% on CIFAR-100). 
For Vision Transformers, however, this trend is not consistent: Parallel ViT achieves higher probe accuracy on CIFAR-100 (88.71\% vs.\ 79.84\%), suggesting that representation learning dynamics depend on architectural inductive bias.

\noindent\textbf{Representation--Inference Mismatch.}
Strong representations do not necessarily translate into strong generative inference. 
For example, the Sequential CNN on CIFAR-100 achieves the best feature quality (85.13\%) but exhibits extremely poor generative performance (14.65\% single-step accuracy), indicating that representation quality alone does not guaranty a consistent vector field.

\medskip
\noindent\textbf{Inference Stability.}
Across architectures, ensemble aggregation generally provides the most stable generative inference, consistent with classical ensemble learning theory [\cite{dietterich2000ensemble}]. 
Multi-step ODE integration, while sometimes beneficial, can suffer from trajectory drift when the learned vector field is inconsistent, particularly in convolutional models [\cite{chen2018neuralode,lipman2023flow}].

\subsection{Analysis of Representation}
\begin{figure}[t]
    \centering
    \includegraphics[width=\textwidth]{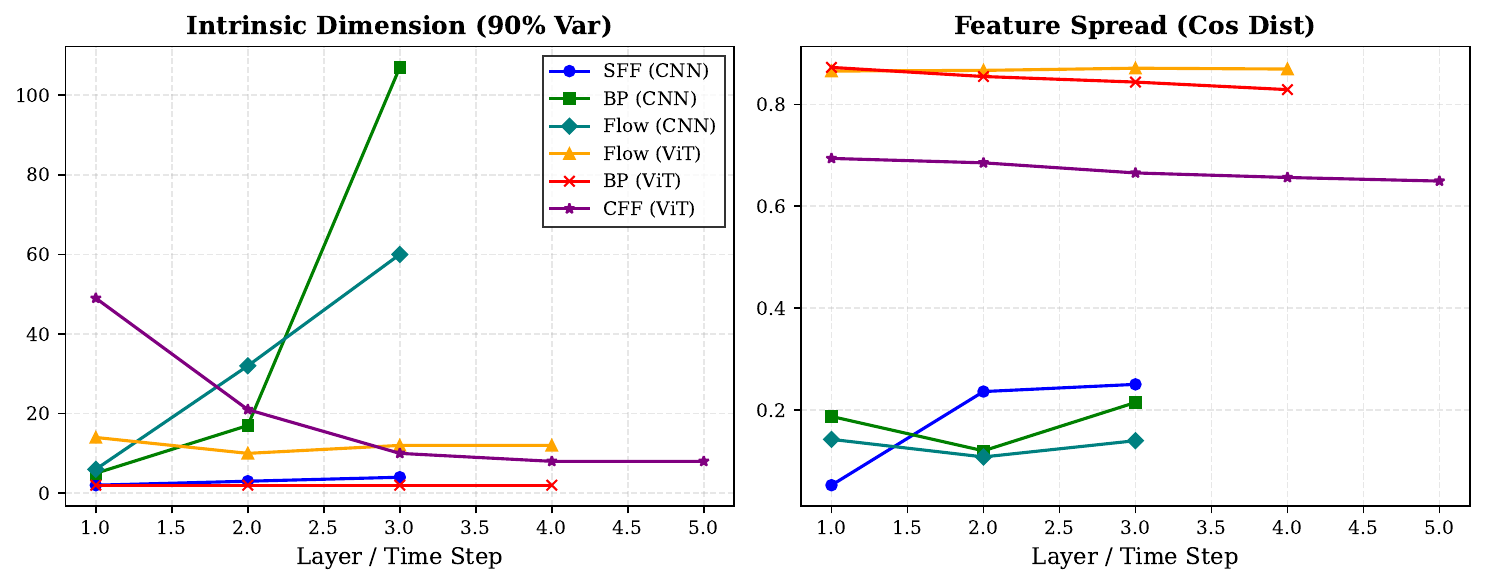}
    \caption{Representation Quality Benchmark comparing Stochastic Forward-Forward (FF), Backpropagation (BP), and Flow Matching (Flow) across CNN and ViT architectures.}
    \label{fig:rep_plot}
\end{figure}

Understanding the geometry of the learned representation space helps reveal
how different optimisation strategies influence model behaviour. To study
these effects, we compare Flow Matching, Backpropagation (BP), and
Forward-Forward variants (CFF and SFF) under matched CNN and Vision
Transformer (ViT) architectures. This controlled comparison isolates the
impact of the training objective while addressing architectural
inductive biases.
We evaluate representations using two complementary metrics. 
\textbf{1. Intrinsic Dimension (ID)} is estimated via PCA at a 90\% variance threshold, where balanced ID reflects efficient capacity use without severe collapse or uncontrolled growth in the representation [\cite{alain2016understanding,raghu2017svcca}]. 
\textbf{2. Feature Spread} is measured as the average pairwise cosine distance between samples, with higher spread indicating more diverse and mathematically distinct features [\cite{li2018intrinsic}].
~\cref{fig:rep_plot} shows the results.

\noindent\textbf{Analysis:}
\begin{itemize}
    \item \textbf{Controlled Manifold Expansion:} The Intrinsic Dimension plot reveals how CNN architectures utilise their expanding channel capacities. Standard BP exhibits representation bloat: as channels increase, BP aggressively spreads features, leading to a sharp increase in intrinsic dimensionality ($>100$). In contrast, Stochastic Forward-Forward (SFF) compresses the manifold by minimising effective dimensionality. However, Flow (CNN), behaves as a geometric regularizer. By constraining representations to follow a continuous generative vector field, the intrinsic manifold expands smoothly and in a controlled manner, capturing features without severe bottlenecks or uncontrolled growth.
    
    \item \textbf{Feature Spread and Inductive Bias:} Feature Spread highlights architectural inductive biases. All CNN methods (BP, FF, and Flow) exhibit relatively low cosine distances, reflecting the tendency of local convolutional receptive fields to produce overlapping features [\cite{zeiler2014visualizing}]. In contrast, Vision Transformers (ViTs) naturally produce highly distinct and orthogonal token representations [\cite{dosovitskiy2020image,touvron2021training}]. Within ViT models, Flow (ViT) maintains highly distinct representations while remaining competitive with standard BP (ViT).
\end{itemize}

CNNs and ViTs differ substantially in their layer-wise dimensional dynamics: CNNs expand channel capacity while reducing spatial resolution, whereas ViTs maintain a constant token dimension. Evaluating Intrinsic Dimension and Feature Spread in pooled channel spaces therefore enables fair comparison without dimensional mismatch. Across these aligned evaluations, Flow Matching consistently acts as a geometric regularizer, producing controlled and structurally coherent manifolds across both backbones.

\section{Conclusion}

In this work, we introduced \textit{Discriminative Flow Matching}, a framework that reformulates discriminative vision tasks as a conditional transport process from noise to task-aligned target manifolds. By attaching multiple independent flow predictors to a shared backbone and training them using local flow-matching objectives, our approach bridges generative continuous-time dynamics with standard discriminative architectures. Experiments on image classification and object detection show that the framework scales across both CNN and Vision Transformer backbones while maintaining competitive performance and reducing training memory through block-wise optimization. Beyond performance evaluation, we analyzed the dynamics of generative inference and the structure of the learned representations. Our results highlight the role of architectural inductive biases on feature quality and generative strength. We also observe that strong feature representations do not always imply consistent vector fields and that ensemble aggregation provides the most stable inference behavior. In general, Discriminative Flow Matching offers a practical bridge between generative transport dynamics and efficient discriminative learning.

\clearpage

\bibliographystyle{unsrtnat}
\bibliography{references}  

\clearpage
\appendix
\renewcommand{\thesection}{\Alph{section}}

\begin{center}
    \Large\bfseries Supplementary Material \\
    \vspace{0.2cm}
    \large Discriminative Flow Matching via Local Generative Predictors
\end{center}
\vspace{0.5cm}
\section{Extended Methods}

\subsection{Local Training Objective}
\label{subsec:local_training}
\subsubsection{Flow Matching Loss}

Each block $k$ is trained independently using a Mean Squared Error loss:
\begin{equation}
\mathcal{L}_{\text{flow}}^{(k)} = \mathbb{E}_{t, z_0, z_1} \left[ \| v_{\theta_k}(z_t, f(x), t) - (z_1 - z_0) \|_2^2 \right]
\label{eq:flow_matching_loss}
\end{equation}

\subsubsection{Task-Specific Anchor Loss}

To ensure the predicted flow is useful for the downstream task, we add an anchor loss. First, we estimate the final state using optimal transport:
\begin{equation}
\tilde{z}_1 = z_t + (1-t)v_{\theta_k}(z_t, f(x), t)
\end{equation}
This estimate is used only for computing task-specific supervision during training
and does not correspond to the inference procedure, which relies on numerical
integration of the learned vector field.

\textbf{For Classification:}
\begin{align}
\ell_{\text{final}} &= W_{\text{cls}}^T \tilde{z}_1 \\
\mathcal{L}_{\text{anchor}}^{(k)} &= -\log \frac{\exp(\ell_{\text{final}}[y])}{\sum_{c=1}^{C} \exp(\ell_{\text{final}}[c])}
\end{align}

\textbf{For Object Detection:} We decode class logits and bounding boxes from $\tilde{z}_1$ and compute a detection loss using Hungarian matching, L1 loss, and Generalized IoU (GIoU):
\begin{equation}
\mathcal{L}_{\text{det}}^{(k)} = \sum_{i=1}^{M} \left[ \mathcal{L}_{\text{cls}}(c_i, \hat{y}_{\hat{\sigma}(i)}) + \mathbf{1}_{\{c_i \neq \emptyset\}} \mathcal{L}_{\text{box}}(b_i, \hat{b}_{\hat{\sigma}(i)}) \right]
\end{equation}

\subsubsection{Combined Loss}

The total loss for each block combines both components:
\begin{equation}
\mathcal{L}_{\text{total}}^{(k)} = \mathcal{L}_{\text{flow}}^{(k)} + \lambda \mathcal{L}_{\text{anchor/det}}^{(k)}
\end{equation}
where $\lambda = 1.0$ in our experiments.

\subsection{Memory Efficiency Analysis}

\subsubsection{Comparison with Standard Backpropagation}

\paragraph{Standard Sequential Architecture.} In a typical deep network with $T$ sequential layers, calculating the gradient requires applying the chain rule to backpropagate the error signal through all layers:
\begin{equation}
\frac{\partial \mathcal{L}}{\partial \theta_1} = \frac{\partial \mathcal{L}}{\partial \hat{y}} \cdot \frac{\partial \hat{y}}{\partial h_T} \cdot \frac{\partial h_T}{\partial h_{T-1}} \cdots \frac{\partial h_2}{\partial h_1} \cdot \frac{\partial h_1}{\partial \theta_1}
\end{equation}

As established in literature [\cite{rumelhart1986learning, chen2016training}], this end-to-end gradient computation suffers from several  bottlenecks:
\begin{itemize}
    \item \textbf{Activation Caching:} It requires storing all intermediate forward activations $\{h_1, \ldots, h_T\}$ in memory to compute the local gradients during the backward pass [\cite{rumelhart1986learning}].
    \item \textbf{Sequential Locking:} The gradients must be computed strictly sequentially from the top layer down to the bottom, which inherently impedes parallel computation.
    \item \textbf{Linear Memory Scaling:} Consequently, the peak activation memory cost scales linearly with network depth, resulting in a memory footprint of $\mathcal{O}(T \cdot B \cdot d)$, where $B$ is the batch size and $d$ is the feature dimension [\cite{chen2016training}].
\end{itemize}
\paragraph{Activation Memory}

To overcome these limitations, our architecture isolates the computational graph for each block:
$$x \xrightarrow{\text{Backbone}} f(x) \rightarrow v_{\theta_k}(z_t, f(x), t) \xrightarrow{\text{Loss}} \mathcal{L}_{\text{total}}^{(k)}$$

By resetting the graph after each local update, we eliminate the sequential dependencies inherent to deep backpropagation. The network only needs to store the forward activations for a single block at any given time, strictly capping the activation memory cost at $\mathcal{O}(B \cdot d)$. 

This decoupling fundamentally alters the dynamics of memory scaling. While the total memory footprint still exhibits a slight linear growth—driven solely by the storage of physical parameters and optimiser states (e.g., momentum) for the $T$ independent blocks—the memory required for activations remains completely constant ($\mathcal{O}(1)$) with respect to depth. By decoupling activation memory from the number of blocks $T$, our method allows for significantly deeper generative trajectories without exhausting hardware resources.

\subsection{Inference Strategies for Learned Flow Dynamics}

Our method learns a \emph{continuous-time vector field} that defines a transport
process from a noise distribution to a task-aligned embedding.
Training enforces \emph{local correctness} of this vector field at arbitrary points
in time, while inference approximates the resulting continuous dynamics using a
discrete numerical solver.

\subsubsection{Training-Time Vector Field Estimation}

During training, each block is supervised to predict the instantaneous vector field
of a predefined probability path.
Given a noise sample $z_0 \sim \mathcal{N}(0, I)$ and a target embedding $z_1$, we
sample a time $t \sim \mathcal{U}[0,1]$ and construct the intermediate state:
\begin{equation}
z_t = (1 - t) z_0 + t z_1.
\end{equation}

The corresponding ground-truth vector field is given by the time derivative of this
path:
\begin{equation}
v^*(z_t, t) = \frac{d z_t}{d t} = z_1 - z_0.
\end{equation}

Each block $v_{\theta_k}$ is trained independently to regress this direction using a
flow matching loss:
\begin{equation}
\mathcal{L}_{\mathrm{flow}}^{(k)} =
\mathbb{E}_{z_0, z_1, t}
\left[
\left\|
v_{\theta_k}(z_t, f(x), t) - (z_1 - z_0)
\right\|_2^2
\right].
\end{equation}

This objective enforces \emph{local correctness} of the learned vector field at
arbitrary points along the trajectory, rather than the direct prediction of the final
state.
\subsubsection{Block-wise Vector Field at Inference}

At inference time, the target embedding $z_1$ is unknown. We therefore interpret the
trained flow blocks as approximations to an underlying continuous-time vector field
that transports an initial noise sample toward a task-aligned embedding, following the formulation of conditional flow matching [\cite{lipman2023flow}].

Depending on the inference strategy, the learned blocks can be used either
\emph{independently}, \emph{sequentially}, or \emph{in ensemble}. This flexibility
mirrors prior work in diffusion and flow-based generative models, where different numerical solvers and sampling strategies trade accuracy for computational efficiency
[\cite{song2021score, karras2022elucidating}].

When multi-step inference is employed, we interpret the collection of $T$ trained
blocks as a \emph{piecewise-defined approximation} of a continuous-time vector field
over the interval $t \in [0,1]$. The interval is partitioned into $T$ equal
sub-intervals, and block $k$ is responsible for predicting the vector field over
$\left[\frac{k-1}{T}, \frac{k}{T}\right]$. The resulting vector field is defined as:
\begin{equation}
v_{\theta}(z_t, f(x), t) = v_{\theta_k}(z_t, f(x), t),
\quad \text{where } k = \left\lfloor T \cdot t \right\rfloor + 1.
\end{equation}

This block-wise formulation avoids global backpropagation through time while retaining
the expressive power of continuous flows is conceptually aligned with locally
trained or decoupled learning paradigms [\cite{hinton2022forward, li2025noprop}.

\subsubsection{Inference via Euler Integration}

\paragraph{Multi-step ODE Inference.}
To recover the final embedding, we numerically approximate the induced ordinary
differential equation using Explicit Euler integration, a standard choice in neural
ODEs and diffusion-inspired models [\cite{chen2018neuralode}]. Starting from an initial
noise sample $z_0 \sim \mathcal{N}(0, I)$, the latent state is updated as:
\begin{equation}
z_{k+1} = z_k + \Delta t \cdot v_{\theta_k}(z_k, f(x), t_k),
\quad \Delta t = \frac{1}{T},
\end{equation}
where $t_k = k \cdot \Delta t$ and $k = 0,1,\ldots,T-1$.

This inference strategy follows the numerical integration procedures commonly used in
continuous normalising flows and score-based generative models
[\cite{grathwohl2019ffjord, song2021score}]. In our framework, multi-step inference is
primarily used for object detection, where iterative refinement of bounding boxes is
beneficial [\cite{carion2020end}].

\paragraph{Single-step Euler Inference.}
In addition to multi-step integration, we employ a simplified single-step inference
strategy. Recent work has shown that well-trained generative vector fields often permit
accurate prediction using a single Euler step
[\cite{lipman2023flow}]. A single block $k$ is queried at $t=0$,
and the final embedding is estimated as:
\begin{equation}
\hat{z}_1 = z_0 + v_{\theta_k}(z_0, f(x), 0).
\end{equation}

This strategy enables extremely fast inference and is particularly effective for
classification tasks, where the target manifold is low-dimensional and globally
structured.

\paragraph{Ensemble Inference (Consensus Transport).}
To further improve robustness, we adopt an ensemble inference strategy inspired by
classical ensembling in discriminative models [\cite{lakshminarayanan2017simple}] and
recent generative ensemble techniques [\cite{ho2022classifierfree}]. All $T$ blocks
independently predict a transport direction from the same initial noise sample:
\begin{equation}
\hat{z}_1^{(k)} = z_0 + v_{\theta_k}(z_0, f(x), 0), \quad k = 1,\ldots,T.
\end{equation}
The final embedding is obtained by averaging these estimates:
\begin{equation}
\bar{z}_1 = \frac{1}{T} \sum_{k=1}^{T} \hat{z}_1^{(k)}.
\end{equation}

Unlike traditional ensembles that require training multiple independent networks, our
approach shares a common backbone and aggregates only the generative transport
predictions, yielding a strong regularisation effect at minimal additional cost.

\subsubsection{Final Prediction}

\paragraph{For Classification.}
The final latent embedding $z$ (obtained via single-step, multi-step, or ensemble
inference) is projected to class logits:
\begin{equation}
\ell = W_{\mathrm{cls}}^{\top} z,
\end{equation}
and the predicted label is:
\begin{equation}
\hat{y} = \arg\max_{c \in \{1,\ldots,C\}} \ell[c].
\end{equation}
This corresponds to nearest-neighbour classification in the learned class embedding
space, a common choice in metric learning and generative classification models
[\cite{snell2017prototypical}].

\paragraph{For Object Detection.}
For object detection, the final query embeddings are decoded into class logits and
bounding boxes using linear prediction heads. Final detections are obtained by
applying non-maximum suppression to the predicted bounding boxes, following standard
practice in detection pipelines [\cite{ren2015faster, carion2020end}].

\subsubsection{Auxiliary Linear Probe for Representation Stabilisation}

In addition to the generative flow matching objective, we employ an auxiliary
\emph{linear probe} on the shared backbone features during training. Given image
features $f(x)$ extracted by the backbone, a linear classifier
$W_{\mathrm{probe}}$ is trained to predict the ground-truth label:
\begin{equation}
\ell_{\mathrm{probe}} = W_{\mathrm{probe}}^{\top} f(x).
\end{equation}

The corresponding cross-entropy loss is defined as:
\begin{equation}
\mathcal{L}_{\mathrm{probe}} =
-\log \frac{\exp(\ell_{\mathrm{probe}}[y])}
{\sum_{c=1}^{C} \exp(\ell_{\mathrm{probe}}[c])}.
\end{equation}

The probe operates on \emph{detached} backbone features, such that gradients
from $\mathcal{L}_{\mathrm{probe}}$ do not propagate into the flow matching blocks.
This auxiliary objective is used solely during training and is discarded at inference
time.

The motivation for this probe is twofold. First, it stabilises representation learning
in the backbone by preventing degenerate solutions when training is driven primarily
by local or decoupled generative objectives. Second, it encourages the learned features
to remain linearly separable, a property known to correlate strongly with downstream
classification performance [\cite{alain2016understanding, he2020momentum}]. Similar
linear evaluation and online probing strategies have been widely used to assess and
stabilise representations in self-supervised and locally trained networks
[\cite{hjelm2019learning,grill2020byol}].

\section{Extended Propositions}

\subsection{Convergence Analysis}

\textbf{Proposition 1 (Local Consistency of Euler Integration).}
Let $v_\theta(z,t)$ denote the learned vector field obtained via flow matching, and
$v^*(z,t)$ the ground-truth vector field induced by the linear interpolation path
between the initial noise $z_0$ and the target embedding $z_1^*$.
Assume that $v_\theta$ is $L$-Lipschitz in $z$ and satisfies the local flow-matching
error bound
\begin{equation}
\mathbb{E}_{z_0,z_1,t}\!\left[\|v_\theta(z_t,t)-v^*(z_t,t)\|_2^2\right] \le \epsilon .
\end{equation}
Then, for a sufficiently small Euler step size $\Delta t$, the discrete Euler update
\begin{equation}
z_{i+1} = z_i + \Delta t\, v_\theta(z_i,t_i)
\end{equation}
satisfies the following local consistency bound:
\begin{equation}
\mathbb{E}\!\left[\|z_{i+1}-z_1^*\|_2^2\right]
\le
\mathbb{E}\!\left[\|z_i-z_1^*\|_2^2\right]
+ \mathcal{O}(\Delta t\,\epsilon)
+ \mathcal{O}(\Delta t^2).
\end{equation}

\begin{figure}[t]
    \centering
    \includegraphics[width=\linewidth]{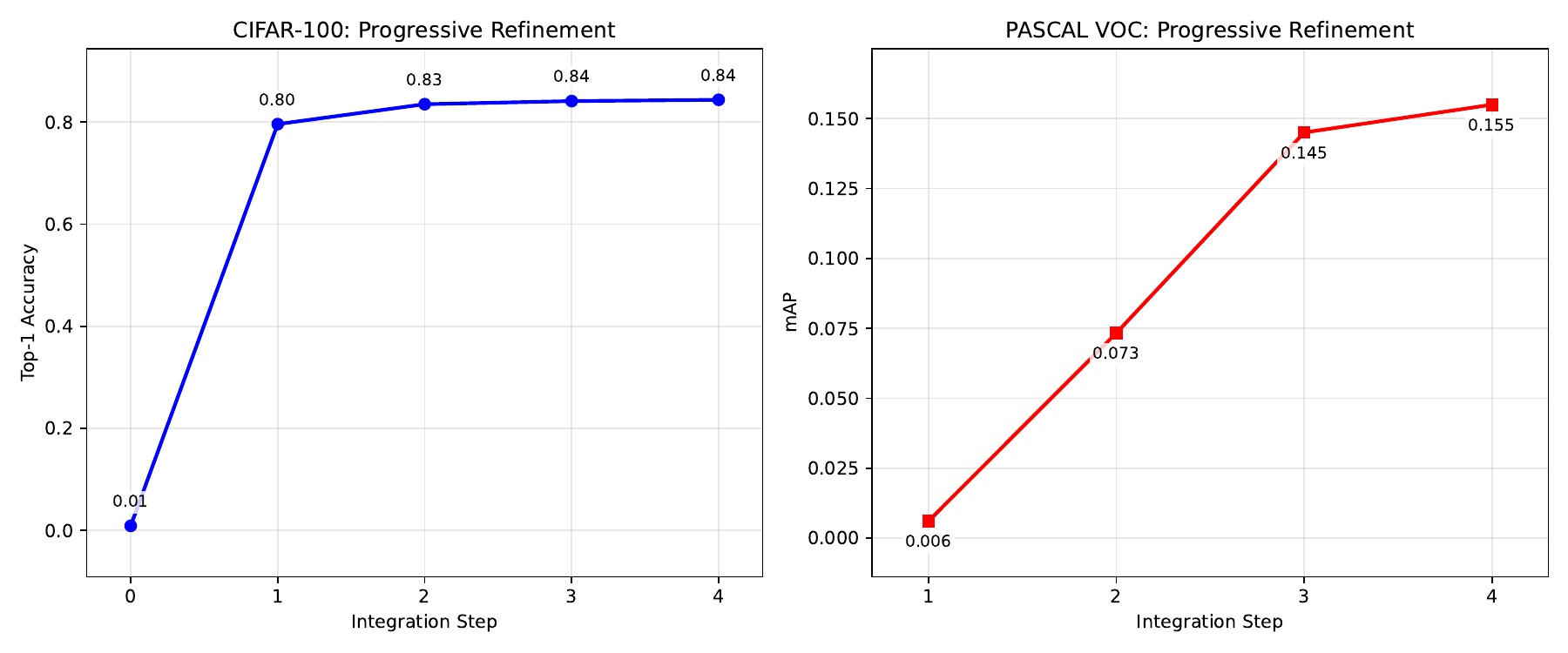}
    \caption{
    \textbf{Progressive task refinement under Euler integration (Proposition~1 \& 2 validation).}
    \emph{Left:} CIFAR-100 classification accuracy evaluated after each integration step. 
    Accuracy improves monotonically and closely matches the behaviour observed during final inference.
    \emph{Right:} PASCAL VOC detection mAP evaluated after each integration step using intermediate decoding of the latent state. 
    Although the absolute mAP values are lower than those obtained under the full detection evaluation pipeline, 
    performance increases consistently across integration steps, indicating progressively more task-aligned latent representations.
    }
    \label{fig:prop1_refinement}
\end{figure}

\paragraph{Interpretation.}
Proposition~1 does not assert global convergence to the target embedding
nor direct minimization of the terminal error $\|z_T - z_1^*\|_2^2$.
Instead, it establishes that each Euler integration step is a locally consistent
approximation of the ideal transport dynamics, with deviations controlled by
the flow-matching error $\epsilon$ and higher-order discretization effects.
As the learned vector field approaches the ground-truth flow ($\epsilon \to 0$),
the numerical error introduced by discretisation remains stable and non-divergent.

\paragraph{Empirical validation.}
To empirically validate this local consistency property, we perform a diagnostic
evaluation in which intermediate latent states are decoded after each integration
step, rather than only at the final inference output.
Figure~\ref{fig:prop1_refinement} illustrates the resulting progressive refinement
behaviour on CIFAR-100 and PASCAL VOC.

On CIFAR-100, we observe a monotonic increase in classification accuracy across
integration steps that closely matches the behaviour of standard inference.
This alignment is expected, as classification depends directly on the final latent
representation and does not involve additional structured decoding stages.

On PASCAL VOC, intermediate decoding yields lower absolute mAP values than full
inference, since object detection performance depends on global query assignment,
bounding-box regression, and post-processing effects such as non-maximum suppression,
which are only fully realised at the final integration step.
Nevertheless, the detection mAP increases consistently across integration steps,
indicating that the latent trajectories become progressively more task-aligned.

Together, these results confirm that the learned vector field induces stable,
non-divergent trajectory refinement under discretised Euler integration,
in agreement with the local consistency guarantee established by Proposition~1.

\subsection{Trajectory Refinement Analysis}

\textbf{Proposition 2 (Progressive Trajectory Refinement).}
Let the continuous-time vector field $v^*(z,t)$ be approximated by a sequence of
$T$ learned vector fields $\{v_{\theta_k}\}_{k=1}^T$, each applied over a disjoint
time interval $[t_k, t_{k+1}]$ with step size $\Delta t = 1/T$.
Assume that each block $v_{\theta_k}$ satisfies a local approximation error
\[
\mathbb{E}\!\left[\|v_{\theta_k}(z,t) - v^*(z,t)\|_2^2\right] \le \epsilon_k,
\quad \forall t \in [t_k, t_{k+1}].
\]
Then the accumulated integration error after $K \le T$ steps satisfies
\[
\mathbb{E}\!\left[\|z_K - z_K^*\|_2^2\right]
\le
\sum_{k=1}^K \mathcal{O}(\Delta t \, \epsilon_k) + \mathcal{O}(K \Delta t^2),
\]
where $z_K^*$ denotes the ideal continuous-time trajectory evaluated at time $t_K$.
In particular, if the local errors $\epsilon_k$ decrease across blocks,
the discrete trajectory exhibits monotonic refinement toward the task-aligned manifold.

\paragraph{Empirical Validation.}
To empirically validate Proposition 2, we refer back to Figure~\ref{fig:prop1_refinement}. As shown, the intermediate representations $z_K$ generated by our model exhibit monotonic improvement in task performance as the trajectory evolves. This confirms that sequential blocks progressively refine the latent representation toward the target manifold, in direct agreement with the accumulated error bounds of Proposition 2.

\subsection{Training Stability Guarantees}
\paragraph{Proof.}
By linearity of differentiation,
\[
\nabla_{\theta_k}\mathcal{L}^{(k)}
=
\nabla_{\theta_k}\mathcal{L}_{\mathrm{flow}}^{(k)}
+
\lambda \, \nabla_{\theta_k}\mathcal{L}_{\mathrm{task}}^{(k)}
+
\beta \, \nabla_{\phi}\mathcal{L}_{\mathrm{shared}} .
\]
Applying the law of total variance yields
{\scriptsize
\begin{align}
\mathrm{Var}\!\left[\nabla_{\theta_k}\mathcal{L}^{(k)}\right]
&=
\mathrm{Var}\!\left[\nabla_{\theta_k}\mathcal{L}_{\mathrm{flow}}^{(k)}\right]
+
\lambda^2 \mathrm{Var}\!\left[\nabla_{\theta_k}\mathcal{L}_{\mathrm{task}}^{(k)}\right] \nonumber\\
&\quad
+ 2\lambda \, \mathrm{Cov}\!\left(
\nabla_{\theta_k}\mathcal{L}_{\mathrm{flow}}^{(k)},
\nabla_{\theta_k}\mathcal{L}_{\mathrm{task}}^{(k)}
\right)
+ \beta^2 \mathrm{Var}\!\left[\nabla_{\phi}\mathcal{L}_{\mathrm{shared}}\right].
\end{align}
}

Bounding the covariance term via the Cauchy–Schwarz inequality yields the stated
upper bound.

Finally, since each block’s forward and backward graphs are constructed and freed
independently during training, neither activation storage nor gradient variance
accumulates with depth $T$, ensuring stable optimization under shared-backbone
training. In particular, in our implementation for Pascal (which we used for this experiment), the shared backbone is initialised from
ImageNet-pretrained weights and fine-tuned using AdamW with a learning rate of
$5\times10^{-4}$ and Layer Normalisation (without Batch Normalisation in decoder blocks). 

\paragraph{Empirical Validation.}
We empirically validate Proposition~3 by measuring the variance of parameter gradients
at each block during training, using identical inputs and pretrained ResNet-50 backbones.
For flow matching, gradients are computed independently for each block using the local
flow-matching objective, while for standard backpropagation, gradients are obtained from
a single end-to-end loss.  Because Flow Matching computes gradients locally, the gradient variance is fundamentally decoupled from the total network depth $T$, preventing the exponential decay inherent to the chain rule.

\begin{figure}[h]
    \centering
    \includegraphics[width=0.8\linewidth]{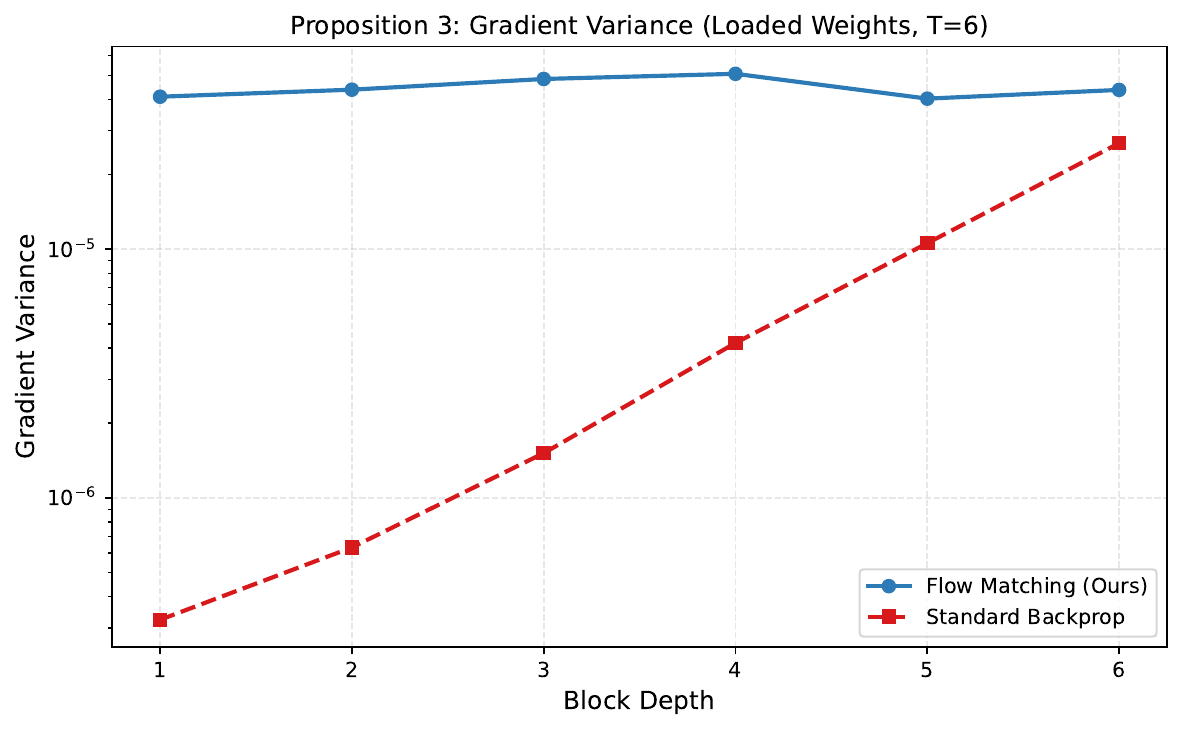}
    \caption{\textbf{Gradient Stability Analysis.} 
    Empirical comparison of gradient variance across blocks ($T=6$) for Flow Matching and Standard Backpropagation. 
    \textbf{Blue (Ours):} Flow Matching maintains constant gradient variance across all depths, validating Proposition 3. 
    \textbf{Orange (Standard):} Standard backpropagation exhibits significantly lower variance in earlier blocks (vanishing gradients), indicating instability in training deep transformer stacks.}
    \label{fig:gradient_stability}
\end{figure}

Figure~\ref{fig:gradient_stability} shows that gradient variance under flow matching
remains stable across depth, whereas standard backpropagation exhibits increasing
variance decay for deeper layers.
Although absolute gradient magnitudes differ, the key observation is the absence of
depth-induced variance attenuation under flow matching.
These results confirm that decoupled block-wise training yields a stable gradient
statistics, even when a backbone is shared.

The advantage of Flow Matching becomes more pronounced as network depth increases. 
In the standard backpropagation baseline (orange), the gradient variance at the first block ($k=1$) decays exponentially with depth $T$. Our experiments showed that increasing $T$ from 4 to 6 reduced the first-layer gradient signal by nearly $10\times$, illustrating the classic vanishing gradient problem that limits deep Transformer training.
In contrast, our Flow Matching approach (blue) exhibits approximately constant variance across depths. The first layer in the $T=6$ model retains $\sim 94\%$ of the variance found in the final layer, compared to only $\sim 1\%$ retention for the standard baseline. This empirically validates Proposition 3 and proves that our method can scale to deeper architectures without optimisation instability.
The gradient variance is evaluated exclusively on decoder and flow-matching blocks, which utilise only Layer Normalisation. Consequently, the observed gradient stability is attributable to decoupled gradient paths rather than normalisation effects.

\subsection{Memory Complexity Analysis}

\textbf{Proposition 4 (Constant Activation Memory).}
\emph{Under blockwise local training, the peak activation memory of our method is
$\mathcal{O}(B \cdot d)$ and is independent of the number of blocks $T$, where $B$
denotes the batch size, and $d$ denotes the latent dimension.}

\paragraph{Standard Backpropagation.}
In conventional deep decoders trained end-to-end, activations from all $T$ blocks
must be retained until the final backward pass:
\begin{equation}
M_{\text{BP}}(T)
=
\mathcal{O}(B \cdot d_{\text{backbone}})
+
\mathcal{O}(T \cdot B \cdot d).
\end{equation}

\paragraph{Our Method.}
In contrast, our training procedure optimises each block independently.
At any iteration, only the activations of the current block are stored:
\begin{equation}
M_{\text{ours}}(T)
=
\mathcal{O}(B \cdot d_{\text{backbone}})
+
\mathcal{O}(B \cdot d).
\end{equation}

During training of block $k$, we:
(i) compute backbone features,
(ii) build the computation graph for block $k$ only,
(iii) perform a local backward pass, and
(iv) immediately discard the graph.
As a result, activation memory does not accumulate with depth.

\paragraph{Asymptotic Advantage.}
Substituting the memory expressions yields
\begin{equation}
\frac{M_{\text{ours}}(T)}{M_{\text{BP}}(T)}
=
\frac{\mathcal{O}(B d_{\text{backbone}}) + \mathcal{O}(B d)}
     {\mathcal{O}(B d_{\text{backbone}}) + \mathcal{O}(T B d)}.
\end{equation}
Since the backbone term is constant with respect to $T$, the denominator grows
linearly while the numerator remains bounded. Consequently,
\begin{equation}
\lim_{T \to \infty}
\frac{M_{\text{ours}}(T)}{M_{\text{BP}}(T)} = 0,
\end{equation}
establishing that our method achieves asymptotically vanishing relative activation
memory compared to standard end-to-end backpropagation.

\section{Extended Experiments}

\subsection{Training from Scratch: CNN and ViT }

To investigate the method's capability to learn representations from scratch without pre-trained weights, we trained both a Convolutional Neural Network (CNN) and a Vision Transformer (ViT) using blockwise local Flow Matching. For CNN experiments, Flow Matching is implemented using blockwise local vector field predictors conditioned on convolutional features, rather than a globally parameterised continuous-time flow as used in the ViT setting.

For the CNN experiments, we utilised an architecture consisting of three convolutional blocks. For the ViT experiments, we used a ViT-Tiny backbone. The results are summarised in Tables \ref{tab: vit_results} and \ref{tab:cnn_results}. The reported accuracy is top 1 per cent for MNIST and CIFAR-10, top 5 per cent for CIFAR-100, and top 10 per cent for Tiny-Imagenet.
\begin{table}[h]

\centering
\begin{tabular}{|l|l|l|l|l|}
\hline
Method & Arch & cifar10 & cifar100 & T\_imagenet \\ \hline
BP & ViT & 81.49 & 78.12  & 74.01 \\ \hline
FF & ViT & 76.21 & 72.15 & 62.11 \\ \hline
CFF+M & ViT & 80.42 & 80.39 & 73.23 \\ \hline
Symba & ViT & 69.13 & 68.78 & 59.46 \\ \hline
FF.Collab & ViT & 70.66 & 71.79 & 62.15 \\ \hline
Ours & ViT & 80.10 & 80.67 & 55.44 \\ \hline
\end{tabular}
\caption{Comparison of ViT architectures on different datasets.}
\label{tab: vit_results}
\end{table}
\begin{table}[h]

\centering
\begin{tabular}{|l|l|l|l|l|}
\hline
Method & Arch & mnist & cifar10 & cifar100 \\ \hline
BP & CNN & 99.33 & 82.5 & 61.28 \\ \hline
FF & CNN & 98.73 & 59 & --- \\ \hline
SFF & CNN & 99.31 & 76.96 & 53.29 \\ \hline
S.Hebb & CNN & 99.35 & 80.31 & 56 \\ \hline
NoProp & CNN & 99.29 & 80.54 & 47.8 \\ \hline
Our & CNN & 98.88 & 78.56 & 59.20 \\ \hline
\end{tabular}
\caption{Comparison of CNN architectures on different datasets.}
\label{tab:cnn_results}
\end{table}
We emphasise that, while our block-wise local Flow Matching objective is consistently applied across both architectures, the geometric spaces of the conditioning features differ: CNN models condition the independent vector field predictors on spatially preserved local feature maps, whereas ViT models condition them on pooled, global latent embeddings. Comparisons, therefore, reflect the behaviour of representational dynamics under different architectural inductive biases rather than strict structural equivalence.

\section{Implementation Details}

\subsection{Object Detection on PASCAL VOC}

For object detection, we evaluate our method on the PASCAL VOC 2007 dataset. Images are resized to $256 \times 256$ and normalised using standard ImageNet statistics. 

\textbf{Architecture:} Both the Flow Matching and Standard Backpropagation models utilise a ResNet-50 backbone pre-trained on ImageNet. The backbone features are projected into an embedding dimension of $d = 128$. The transformer-based blocks (for both vector field prediction and standard decoding) use 4 attention heads. We deploy $M = 20$ object queries to detect up to 20 classes. For our memory studies, we vary the number of blocks/integration steps $T \in \{4, 8, 16, 20, 24\}$.

\textbf{Training Configuration:} Models are trained for 50 epochs with a batch size of 16. We use the AdamW optimiser with a weight decay of $10^{-4}$. The base learning rate for the flow matching blocks is set to $10^{-4}$, while the pre-trained backbone is fine-tuned with a lower learning rate of $10^{-5}$. We apply a step learning rate decay, reducing the learning rates by a factor of 0.1 at epoch 20. Gradients are clipped to a maximum norm of 1.0.

\textbf{Loss Formulation:} Bounding box assignments are determined via bipartite Hungarian matching. The matching costs and training loss weights are strictly aligned: $\lambda_{\text{cls}} = 1.0$ for cross-entropy classification, $\lambda_{\text{bbox}} = 5.0$ for $L_1$ bounding box regression, and $\lambda_{\text{giou}} = 2.0$ for Generalised IoU loss. The background class ("no object") is scaled by a factor of 0.1. For our method, the Flow Matching MSE loss is weighted by $0.1$ relative to the detection losses.

\textbf{Inference:} During inference, the Flow Matching model generates proposals via Euler integration over $T$ steps ($dt = 1/T$). Predictions are filtered using a confidence threshold of 0.05 before evaluating the mean Average Precision (mAP) at an IoU threshold of 0.5.
\subsection{Image Classification}

To comprehensively assess our framework, we evaluate image classification performance across CIFAR-10, CIFAR-100, TinyImageNet, and ImageNet-R.

\textbf{Architectures (Training from Scratch):} We train Vision Transformers (ViTs) and Convolutional Neural Networks (CNNs) from scratch without pre-trained weights, carefully matching capacities to established baselines. 
\begin{itemize}
    \item \textbf{ViT (CIFAR):} Operating on $32 \times 32$ images ($4 \times 4$ patches). CIFAR-100 utilizes a ViT[192 6 6] configuration ($d=192$, depth 6, 6 heads) with $T=6$ flow blocks. CIFAR-10 utilizes ViT[128 4 5] ($d=128$, depth 5, 4 heads) with $T=5$ flow blocks. Global pooling is omitted to preserve the spatial token sequence.
    \item \textbf{ViT (TinyImageNet):} Operating on $64 \times 64$ images ($8 \times 8$ patches). We use a ViT[240 6 7] configuration with $T=7$ flow blocks. To improve early optimisation stability, the standard linear patch projection is replaced with an $8 \times 8$ convolutional stem. The flow predictors utilise GELU activations and accept concatenated inputs of the latent state, time embedding, and spatial memory $[z_t, t_{emb}, f(x)]$. 
    \item \textbf{CNN (CIFAR-10 and CIFAR-100):} The backbone consists of three sequential convolutional blocks (96, 384, and 1536 channels). We evaluate two distinct Flow Matching topologies: \textit{Parallel Consensus} (where independent predictors attach to the shared final representation) and \textit{Sequential} (where predictors attach block-by-block). The flow predictors operate on a latent dimension of $d=512$. While most CNN configurations utilise deep residual MLPs (hidden dimension 1024), we replace these with lighter, single-hidden-layer MLPs (with Dropout) for the CIFAR-100 Sequential setting to mitigate overfitting. 
\end{itemize}
In our ViT parallel consensus experiments, flow predictors are upgraded to deeper residual MLPs (hidden dimension 1024 with LayerNorm and Dropout) to increase local capacity.

\textbf{Architectures (Pre-trained Fine-Tuning):} For experiments on CIFAR-10, CIFAR-100, and ImageNet-R, we utilize a ViT-B/16 backbone pre-trained on ImageNet-1K ($d=768$). The original classification head is removed, and $T=4$ independent local flow predictors are attached to the unpooled output sequence.

\textbf{Training Configurations:} We apply distinct optimisation strategies depending on the initialisation paradigm. Gradients are clipped to a maximum norm of 1.0.
\begin{itemize}
    \item \textbf{From Scratch:} Models are optimised using AdamW with a base learning rate of $5 \times 10^{-4}$ (applied equally to the backbone and flow heads) and weight decay of $10^{-4}$. The schedule uses a 5-epoch linear warmup followed by cosine annealing. Epochs and batch sizes scale by dataset difficulty: CIFAR-100 ViT (300 epochs, BS 512), CIFAR-10 ViT (50 epochs, BS 128), CNNs (200 epochs, BS 128, EMA decay 0.995), and TinyImageNet (1500 epochs, BS 256).
    \item \textbf{Fine-Tuning:} We apply differential learning rates to maintain pre-trained backbone stability while training the generative flow heads. The objective jointly optimises the optimal transport Flow Matching MSE loss alongside a standard cross-entropy loss evaluated on the $t=1$ estimate. CIFAR-10 and CIFAR-100 are trained for 20 and 10 epochs, respectively (BS 32, Backbone LR $1 \times 10^{-6}$, Head LR $1 \times 10^{-4}$). ImageNet-R is trained for 50 epochs (BS 64, Backbone LR $1 \times 10^{-6}$, Head LR $1 \times 10^{-3}$).
\end{itemize}

\textbf{Baseline Configurations (Standard Backpropagation):} To ensure fair methodological comparisons, standard backpropagation baselines (denoted as BP) are evaluated using controlled environments that isolate our local objective from architectural capacity. 
\begin{itemize}
    \item \textbf{Fine-Tuning Baselines:} For CIFAR-10, CIFAR-100, and ImageNet-R, the BP baselines use the identical ViT-B/16 backbone. To exactly match the parameter capacity and depth of our $T=4$ flow blocks, the baseline classification head is replaced with a deep stack of 4 standard Transformer Encoder Layers. These baselines are trained with identical hyperparameters (batch sizes, epochs, data augmentations, and differential learning rates) to those of our corresponding fine-tuning models.
    \item \textbf{From-Scratch Baselines:} The from-scratch ViT and CNN baselines mirror the structural complexities of their respective Flow Matching counterparts but are trained end-to-end for 60 epochs (Batch Size 128). They utilise AdamW with Cosine Annealing, a weight decay of $0.01$, and learning rates of $5 \times 10^{-4}$ (ViT) and $1 \times 10^{-3}$ (CNN).
\end{itemize}

\textbf{Data Augmentation \& Evaluation:} To mitigate overfitting on harder tasks from scratch, CIFAR-100 (ViT and CNN), CIFAR-10 (CNN), and TinyImageNet apply AutoAugment. CIFAR-100 and TinyImageNet additionally apply Random Erasing ($p=0.25$), with TinyImageNet also including Random Rotation ($20^\circ$). CIFAR-10 (ViT) utilises standard random cropping and horizontal flipping. For from-scratch representation learning, final classification accuracy is also evaluated via a linear probe trained on the frozen backbone features, alongside generative ensemble inference and multi-step ODE solvers.

\subsection{Hardware and Software Environment}

\textbf{Hardware Configuration:}
\begin{itemize}
    \item GPU: NVIDIA RTX 4500 Ada Generation (24GB VRAM)
    \item CPU: Intel Core i9-14900K
    \item RAM: 125GB DDR4
\end{itemize}

\textbf{Software Stack:}
\begin{itemize}
    \item PyTorch 2.5.1 with CUDA 12.1
    \item Python 3.10.18
    \item torchvision 0.20.1
    \item NumPy 2.0.1
    \item SciPy 1.15.3
    \item Matplotlib 3.10.5
\end{itemize}





\end{document}